\documentclass[runningheads]{llncs}
\usepackage{graphicx}
\usepackage{tikz}
\usepackage{comment}
\usepackage{amsmath,amssymb} % define this before the line numbering.
\usepackage{color}

\usepackage[accsupp]{axessibility}  % Improves PDF readability for those with disabilities.

\usepackage[width=122mm,left=12mm,paperwidth=146mm,height=193mm,top=12mm,paperheight=217mm]{geometry}

\usepackage{bm}
\usepackage{multirow}
\usepackage{makecell}
\usepackage{booktabs}
\usepackage[pagebackref,breaklinks,colorlinks]{hyperref}
\usepackage{subfigure}

\begin{document}
% \linenumbers
\pagestyle{headings}
\mainmatter

\makeatletter
\newcommand{\printfnsymbol}[1]{%
	\textsuperscript{\@fnsymbol{#1}}%
}
\makeatother

\title{Generative Subgraph Contrast for Self-Supervised Graph Representation Learning} % Replace with your title

%\begin{comment}
\titlerunning{Generative Subgraph Contrast}

\author{Yuehui Han\and Le Hui\and Haobo Jiang\and Jianjun Qian\thanks{Corresponding authors}\and Jin Xie\printfnsymbol{1}}
\authorrunning{Y.Han\and L.Hui\and H.Jiang\and J.Qian\and J.Xie}

\institute{Key Lab of Intelligent Perception and Systems for High-Dimensional Information of Ministry of Education \\ 
	Jiangsu Key Lab of Image and Video Understanding for Social Security \\
	PCA Lab, School of Computer Science and Engineering \\ Nanjing University of Science and Technology, China \\
\email{\{hanyh, le.hui, jiang.hao.bo, csjqian, csjxie\}@njust.edu.cn} }
%\end{comment}
%******************
\maketitle

\begin{abstract}
Contrastive learning has shown great promise in the field of graph representation learning. By manually constructing positive/negative samples, most graph contrastive learning methods rely on the vector inner product based similarity metric to distinguish the samples for graph representation. However, the handcrafted sample construction (e.g., the perturbation on the nodes or edges of the graph) may not effectively capture the intrinsic local structures of the graph. Also, the vector inner product based similarity metric cannot fully exploit the local structures of the graph to characterize the graph difference well. To this end, in this paper, we propose a novel adaptive subgraph generation based contrastive learning framework for efficient and robust self-supervised graph representation learning, and the optimal transport distance is utilized as the similarity metric between the subgraphs. It aims to generate contrastive samples by capturing the intrinsic structures of the graph and distinguish the samples based on the features and structures of subgraphs simultaneously. 
Specifically, for each center node, by adaptively learning relation weights to the nodes of the corresponding neighborhood, we first develop a network to generate the interpolated subgraph. We then construct the positive and negative pairs of subgraphs from the same and different nodes, respectively. Finally, we employ two types of optimal transport distances (i.e., Wasserstein distance and Gromov-Wasserstein distance) to construct the structured contrastive loss. 
Extensive node classification experiments on benchmark datasets verify the effectiveness of our graph contrastive learning method. Source code is available at \url{https://github.com/yh-han/GSC.git}.

%We would like to encourage you to list your keywords within the abstract section
\end{abstract}

\section{Introduction}

Graph representation learning \cite{hamilton2017representation} has received intensive attention in recent years due to its superior performance in various downstream tasks, such as node/graph classification \cite{kipf2016semi,lee2019self}, link prediction \cite{zhang2018link} and graph alignment \cite{faerman2019graph}. 
Most graph representation learning methods \cite{hamilton2017inductive,kipf2016semi,velivckovic2017graph} are supervised, where manually annotated nodes are used as the supervision signal. 
Since the acquisition of supervision signals is time-consuming and labor-intensive, these methods are difficult to be applied to real scenarios.

%\begin{figure}[t]
%	\centering
%	\includegraphics[width=0.98 \linewidth]{fig/introduction/intro_cropped.pdf}
%	\caption{Example of graph data augmentation. Edges deletion based method (a) and edges addition based method (b) may change the original graph structure information by controlling edges to add and delete graph nodes. Based on the complete neighborhood structure, our method (c) is able to construct more effective contrastive samples.}
%	\label{intro}
%\end{figure}

Recent efforts have been devoted to unsupervised graph representation learning \cite{jin2020self,kipf2016variational,manessi2021graph,wu2021self,wu2020comprehensive,zhu2020self}. 
Among these methods, graph contrastive learning is a powerful manner for learning node or graph representation.
By manually constructing positive/negative samples based on the perturbation (e.g., attribute masking, nodes shuffling or edge perturbation), it aims to enforce similar samples to be closer and dissimilar samples far from each other.
However, dropping edges or masking node attributes randomly may change the original properties of the graph.
For example, by adding or discarding graph nodes to construct positive samples, edge perturbation based methods \cite{hafidi2020graphcl,suresh2021adversarial,zhu2020deep,zhu2021graph} may change the local geometric structures of the original graph, resulting in generating dissimilar positive samples. 
Thus, with GNNs for feature extraction, the features of positive samples cannot be guaranteed to be as close as possible in the graph contrastive learning framework. 
Moreover, since the perturbation-based positive/negative sample augmentation methods are dataset-specific, it is difficult to adaptively select the suitable augmentation method for the specific dataset. 
Besides, the readout function is usually used to construct the vector-wise similarity metric between nodes/graphs, which ignores the structures of the graph \cite{hassani2020contrastive,jiao2020sub,velivckovic2018deep}. 
Thus, the vector inner product based similarity metrics cannot characterize the graph difference well.

In this paper, instead of manually constructing contrastive samples, we propose a novel subgraph generation based contrastive learning framework for efficient self-supervised graph representation learning, where the optimal transport distance is employed to capture the difference between the subgraphs for robust similarity evaluation. 
Specifically, we first sample the neighbor subgraph of the center node based on the breadth first search (BFS).
We then develop a subgraph generation network to adaptively generate subgraphs whose nodes are interpolated in the feature space with the learned weights.  
For each node, we can assign different attentional weights to the neighboring nodes to obtain the weighted node so that the formed subgraph can capture the intrinsic geometric structure of the graph. 
Consequently, we construct the positive pair with the sampled subgraph and the generated subgraph of the same center node and the negative pair with the sampled and generated subgraphs of different center nodes.
Finally, based on the constructed positive/negative subgraphs, we formulate the structured contrastive loss to learn the node representation with the Wasserstein distance and Gromov-Wasserstein distance \cite{chen2020graph}. 
The structured contrastive loss can minimize the geometry difference between the positive subgraphs and maximize the difference between the negative subgraphs.
Experimental results on five benchmark node classification datasets demonstrate that our proposed graph contrastive learning method can yield good classification performance.

%To the best of our knowledge, this is the first work employing subgraph generation and optimal transport distance for graph contrastive learning.

To summarize, the main contributions include:
\begin{itemize}
	\item We propose a novel adaptive sample generation based contrastive learning framework for self-supervised graph representation learning. 
	\item We develop a subgraph generation module to adaptively generate contrastive subgraphs with neighborhood interpolation.
	\item We employ the optimal transport distance as the similarity metric for subgraphs, which can distinguish the contrastive samples by fully exploiting the local attributes (i.e., features and structures) of the graph.
\end{itemize}

\section{Related Work}

\subsection{Graph Neural Networks}
The purpose of graph neural networks (GNNs) is to use graph structures and node features to learn the node representations. 
Formally, classical GNNs follow a two-step processing: neighborhood node aggregation and feature transformation. 
It first updates the node representations by aggregating the representations of its neighboring nodes as well as its representations. 
Then, the representations of each node are mapped into a new feature space by the shared linear transformation. 
Graph Convolutional Network (GCN) \cite{kipf2016semi} employs a weighted sum of the 1-hop neighboring node features to update the node features, where the weights of each node come from the node degree. 
Graph Attention Network (GAT) \cite{velivckovic2017graph} calculates the weights by using the interaction between the neighboring nodes to replace the node degree.
However, they usually need the complete graph as the input. Therefore, limited by the hardware resources, these methods are not suitable to be applied to large-scale graph data.
To solve this issue, Hamilton et al \cite{hamilton2017inductive} propose the sampling-based method, GraphSAGE. 
They first sample the neighborhood nodes for the mini-batch of the center nodes and update the node features by aggregating the sampled neighborhood nodes. 
Then, the batch nodes are iteratively updated until the entire graph is updated. 
These methods mainly focus on supervised learning and require a lot of manual labels. 
However, the acquisition of manually annotated labels is costly in labor and time. 

\subsection{Graph Contrastive Learning}
Graph contrastive learning has recently been considered a promising approach for self-supervised graph representation learning. 
Its main objective is to train the encoder with an annotation-free pretext task. 
The trained encoder can transform the data into low-dimensional representations, which can be used for downstream tasks. 
The basic idea of graph contrastive learning aims at embedding positive samples close to each other while pushing away each embedding of the negative samples.  
In general, we can divide graph contrastive learning into two categories: pretext task based and data augmentation based methods.

\textbf{Pretext Task.} In graph contrastive learning, many early works design pretext tasks from the scale of the contrastive samples, i.e., node, subgraph or graph. 
Inspired by Deep InfoMax (DIM) \cite{hjelm2018learning}, Deep Graph Infomax (DGI) \cite{velivckovic2018deep} and Mutual Information Graph (INFOGRAPH) \cite{sun2019infograph} learn the representations of nodes or graph by maximizing mutual information between the node and global graph. 
Also based on the contrast of nodes and graph, Multi-View Graph Representation Learning (MVGRL) \cite{hassani2020contrastive} expands DGI to multiple views.
By adding the cross-view contrast between the representation of nodes and graph, MVGRL further enhances the guidance performance of the pretext task.
In order to avoid the problem of sharing positive samples (global graph) among multiple nodes in these methods, some works try to construct exclusive positive sample for each sample.
Graphical Mutual Information (GMI) \cite{peng2020graph} proposes to maximize the mutual information between the neighborhood in input and the center node in output.
Sub-graph Contrast (Subg-Con) \cite{jiao2020sub} learns node features by taking the induced subgraphs of the center node as the input of the encoder and treating the center node and context subgraph as the contrastive sample pairs. This method can also alleviate the problem of memory overload caused by large-scale graphs. 
By treating top-k similar nodes from T-hop neighbors as positive samples, Augmentation-Free Graph Contrastive Learning (AF-GCL) \cite{wang2022augmentation} proposes the augmentation-free methods.
Graph Contrastive Coding (GCC) \cite{qiu2020gcc} proposes to contrast between subgraphs. It takes the subgraphs from the same r-ego network as positive samples and subgraphs from the different r-ego networks as negative samples. However, GCC only considers the structure information neglecting the node features. 

%To reduce the labeling effort, the GPT-GNN \cite{yu2018generative} is proposed to initialize GNNs by generative pre-training. The generation can be decomposed into two coupled parts: 
%Given the observed edges, GPT-GNN generates node attributes; given the observed edges and generated node attributes, GPT-GNN generates the remaining edges.
%By treating top-k similar nodes from T-hop neighbors as positive samples, Augmentation-Free Graph Contrastive Learning (AF-GCL) \cite{wang2022augmentation} proposes the augmentation-free methods.
%Besides preserving the local similarities of the graph, Local-instance and Global-semantic Learning (GraphLoG) \cite{xu2021self} introduces the hierarchical prototypes to capture the global semantic clusters to learn the whole-graph representation.
%To solve the problem of false-negative samples, \cite{zhao2021graph} proposes to jointly perform representation learning and clustering, where feature representation and clustering can be promoted from each other.
%%Also based on the negative sample selection problem, 
%Curriculum Contrastive Learning (CuCo) \cite{Chu2021CuCoGR} proposes a scoring function to sort the negative samples from easy to hard, and a pacing function to automatically select the negative samples in the training process.
In addition to the scale of the contrastive samples, some works design pretext tasks to better exploit contrastive information.
To solve the problem of false-negative samples, \cite{zhao2021graph} proposes to jointly perform representation learning and clustering, where feature representation and clustering can be promoted from each other.
With the same motivation, Curriculum Contrastive Learning (CuCo) \cite{Chu2021CuCoGR} proposes a scoring function to sort the negative samples from easy to hard, and a pacing function to automatically select the negative samples in the training process.
For better selection of positive samples, Augmentation-Free Graph Representation Learning (AFGRL) \cite{lee2022augmentation} proposes to discover the positive node that share the local structural information and the global semantics.
Besides, Local-instance and Global-semantic Learning (GraphLoG) \cite{xu2021self} proposes to capture the local similarities and the global semantic clusters to learn the whole-graph representation.

\textbf{Data Augmentation.} Data augmentation based graph contrastive learning methods usually design different perturbation manners (e.g., attribute masking, nodes shuffling or edge perturbation) to construct contrastive samples.
Deep Graph Contrastive Representation Learning (GRACE) \cite{zhu2020deep} augments the graph by setting the probability of edge removal and node features mask. Then, it takes the corresponding nodes of the augmented graph as positive samples and all the other nodes as negative samples.
Graph Contrastive Learning (GraphCL) \cite{hafidi2020graphcl} proposes the sample augmentation manner from the subgraph level. For the induced subgraphs of the center nodes, it employs two stochastic perturbations and a shared encoder to produce two representations of the same node. 
You et al propose \cite{you2020graph} for molecular property prediction in chemistry and protein function prediction in biology. They systematically study the effects of various combinations of graph augmentations on multiple datasets, and found that the choice of data augmentation is closely related to the specific datasets.
However, these perturbation-based methods heavily rely on handcraft settings. 
To solve this problem, various efforts have been made.
On the one hand, some works try to optimize the setting of perturbation probability.
Based on the node centrality, Graph Contrastive Learning with Adaptive Augmentation (GCA) \cite{zhu2021graph} can adaptively learn the probability of edge removal. Besides, by adding more noise to the unimportant node features, it can enforce the model to recognize underlying semantic information.
Based on the min-max principle, Adversarial Graph Contrastive Learning (AD-GCL) \cite{suresh2021adversarial} proposes a trainable edge-dropping graph augmentation manner. 
On the other hand, some works try to optimize the choice of perturbation methods.
%Following \cite{you2020graph}, in order to solve the problem of the choice of graph data augmentation manners, 
Joint Augmentation Optimization (JOAO) \cite{you2021graph} proposes to adaptively select data augmentation manners of graph by adversarial training. 
With the same motivation, Automated Graph Contrastive Learning (AutoGCL) \cite{yin2022autogcl} proposes to adaptively select data augmentation manners of nodes by the learnable graph view generator.

%------------------------------------------------------------------------
\section{Proposed Method}
\label{sec:formatting}
\begin{figure*}[t]
	\centering
	\includegraphics[height=0.42 \linewidth]{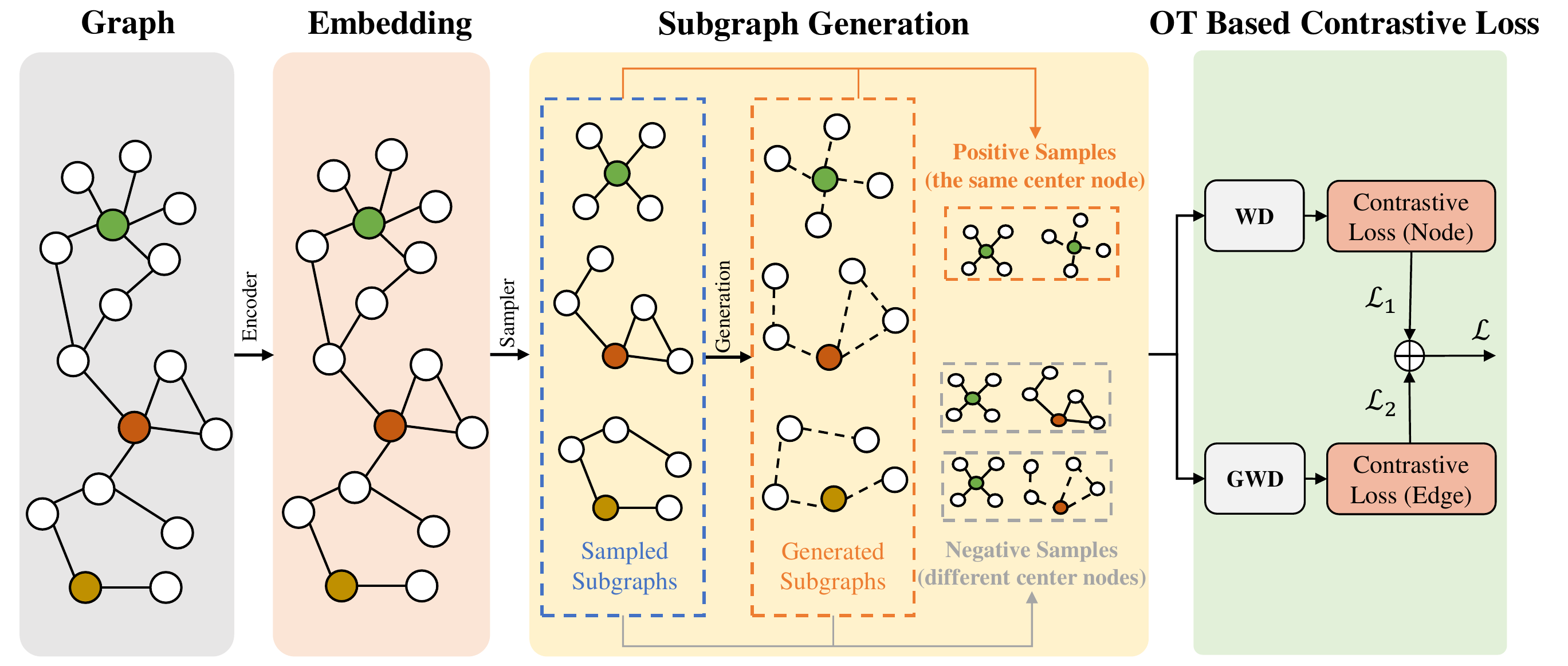}
	\caption{The architecture of our method. We first employ an encoder to obtain the node embeddings. Based on the BFS sampling, we obtain the subgraphs of each node. Next, we use the proposed generation module to generate the contrastive samples of the sampled subgraphs. Then we take the sampled and generated subgraphs with the same center node as the positive samples while the subgraphs with the different center nodes as the negative samples. In order to fully exploit local structure information of the graph, we further introduce two types of optimal transport distances (i.e., Wasserstein distance and Gromov-Wasserstein distance) to calculate the similarity between the subgraphs. Finally, we use the combination of the WD-based contrastive loss $\mathcal{L}_{1}$ and GWD-based contrastive loss $\mathcal{L}_{2}$ to train the network.}
	\label{architecture}
\end{figure*}

In this section, we present our subgraph generation based graph contrastive learning method. As shown in Fig. \hyperref[architecture]{1}, 
based on the sampled subgraphs with the breadth first search, we first adaptively generate the contrastive subgraphs to construct positive/negative samples. Then, we employ the optimal transport distance (i.e., Wasserstein distance and Gromov-Wasserstein distance) to formulate the contrastive loss between the constructed samples.

%-------------------------------------------------------------------------
\subsection{Adaptive Subgraph Generation}
Before introducing our method, we first provide the preliminary concepts about our graph representation learning. 
Let $G=(V, E)$ represent an undirected graph, where $V$ and $E$ denote the vertex set and the edge set, respectively. 
The feature matrix of the graph is denoted as $\bm{X}=\{\bm{x}_1, \bm{x}_2,..., \bm{x}_N \}$, where $\bm{x}_i\in R^{C}$ is the feature of the node $i$, $C$ represents the dimension of input features and $N$ is the number of the nodes.
The adjacency matrix $\bm{A} \in \mathbb{R}^{N \times N}$ indicates the topological structure of the graph where if node $i$ and $j$ are linked, $\bm{A}_{ij}=1$, otherwise, $\bm{A}_{ij}=0$. 
Let $\mathcal{G}_{i}=(V_{i}, E_{i})$ denote the induced subgraph of center node $i$, where $V_{i}$ and $E_{i}$ represent the vertex set and the edge set of the induced subgraph $i$, respectively.  We denote the adjacency matrix of subgraph $i$ induced from graph as $\bm{A}_{i} \in R^{k \times k}$, where $k$ is the number of nodes of subgraph $i$. The goal of self-supervised graph representation learning is to learn the nodes embeddings $\bm{H} = \varepsilon(\bm{A}, \bm{X})$ via an encoder $\varepsilon: R^{N \times C} \times R^{N \times N} \rightarrow R^{N \times F}$ without supervised information, where $F$ is the dimension of embeddings.

The construction of contrastive samples is critical in graph contrastive learning.
Most graph contrastive learning methods generate positive and negative samples with the perturbation of nodes, edges, or graphs.
The perturbation operation may lose important information or even destroy the intrinsic structures of the graph. 
Thus, the constructed samples may be not discriminative enough to train the contrastive learning model. 
In order to construct more effective contrastive samples, we propose a learnable subgraph generation module to generate positive/negative subgraph samples. 
It is expected that the generated subgraphs can characterize the intrinsic local structures of the graph well.

\begin{figure}[t]
	\centering
	\includegraphics[width=0.8 \linewidth]{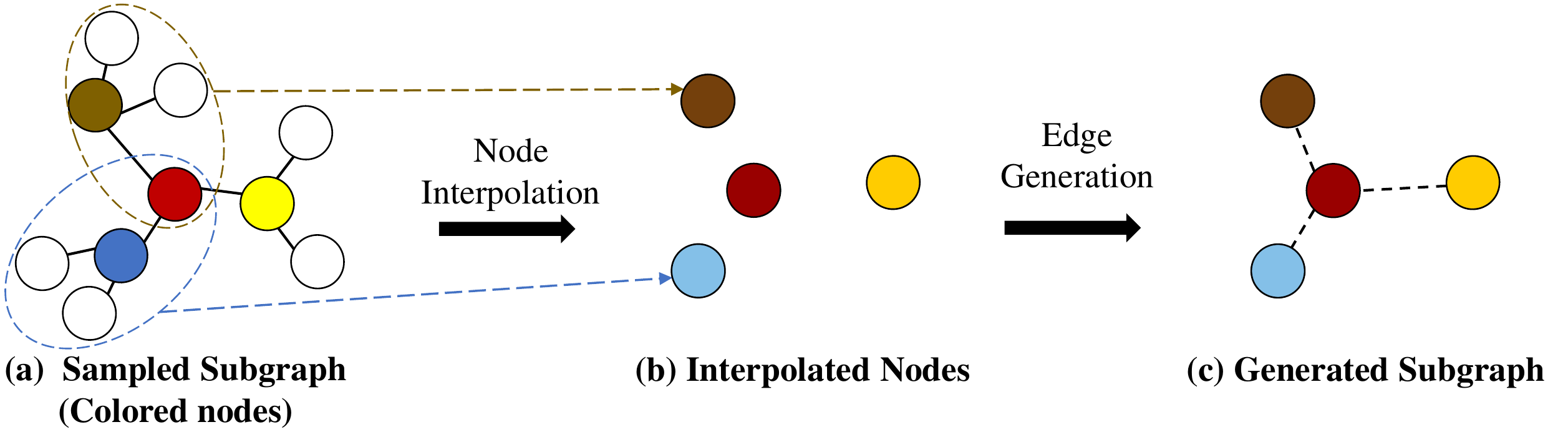}
	\caption{The proposed interpolation-based adaptive subgraph generation module. 
		For the sampled subgraph in (a), we use the neighborhood of each subgraph node to interpolate the new node in the subgraph (b).
		Based on the interpolated nodes features, we then generate the edges between the interpolated nodes. 
		Finally, we obtain the generated subgraph (c).}
	\label{generation}
\end{figure}

The proposed generation module can adaptively generate the contrastive subgraph of the sampled subgraph. (For the specific sampling process, please refer to the supplementary materials).
As shown in Fig. \hyperref[generation]{2} (a)(b), based on the local structure information interpolation, we first generate the subgraph nodes. 
Then, we generate the edges of the subgraph based on the interpolated nodes.

For a specific sampled subgraph node $i$, we can formulate the interpolation-based generation as:
\begin{equation}\label{key}
	\bm{\hat{h}}_{i} = \sum^{\mathcal{N}_{i}}_{j=1} a_{j} \bm{h}_{j}
\end{equation}
where $\bm{h}_{j} \in R^{F}$ is the representations of neighborhood node of center node $i$. $j \in \mathcal{N}_{i}$, $\mathcal{N}_{i}$ is the neighborhood of node $i$ in the graph.
$\bm{\hat{h}}_{i} \in R^{F}$ is the representations of generated subgraph node. $a_{j}$ is the learned relationship weight between the neighborhood node $j$ and the center node $i$.
For each sampled subgraph node, we perform the interpolation based on learned neighborhood relation weights to generate new nodes.
As for the learned relation weight $a_{j}$, we can define it as:
\begin{equation}\label{key}
	a_{j} = \frac{exp(\theta(\bm{h}_{i}, \bm{h}_{j}))}{\sum^{\mathcal{N}_{i}}_{k=1}  exp(\theta(\bm{h}_{i}, \bm{h}_{k})) }
\end{equation}
where $\theta(\bm{h}_{i}, \bm{h}_{j})$ represents the relationship between center node $i$ and neighborhood node $j$. We can define the $\theta(\bm{h}_{i}, \bm{h}_{j})$ as:
\begin{equation}\label{key}
	\theta(\bm{h}_{i}, \bm{h}_{j}) = \operatorname{LeakyReLU}(\bm{W}_{\theta} [\bm{W}_{\phi}\bm{h}_{i} \Vert \bm{W}_{\phi}\bm{h}_{j} ])
\end{equation}
where LeakyReLU is the activation function (with negative input slope 0.2), $\bm{W}_{\theta} \in R^{1\times2F}$ and $\bm{W}_{\phi} \in R^{F \times F}$ are the weight matrixes to be learned, and $\Vert$ represents the feature concatenation.

As shown in Fig. \hyperref[generation]{2} (c), based on the generated nodes, we directly generate the edges of the contrastive subgraph. 
For the node $s_i$ and $s_j$ in the subgraph, $s_i, s_j = 1,2,...,k$, $k$ is the number of subgraph nodes, the generated edge between nodes $s_{i}$ and $s_{j}$ of the subgraph can be denoted as:
\begin{equation}\label{key}
	\bm{\hat{A}}(s_i,s_j) = \varphi(\bm{\hat{h}}_{s_i}, \bm{\hat{h}}_{s_j})
\end{equation}
where $\bm{\hat{A}}$ is the adjacency matrix of generated subgraph, $\bm{\hat{h}}_{s_i}$ is the generated features of subgraph node $s_i$. $\varphi(.,.)$ is the similarity calculation function, here, we use the cosine similarity, i.e., $\varphi(\bm{\hat{h}}_{s_i}, \bm{\hat{h}}_{s_j})=\frac{\bm{\hat{h}}_{s_i}^{T}\bm{\hat{h}}_{s_j}}{\Vert\bm{\hat{h}}_{s_i}\Vert_2 \Vert\bm{\hat{h}}_{s_j}\Vert_2}$.

So far, we obtain the generated contrastive subgraph that contains the nodes features and edges. 
Essentially, we use the adaptively generated samples to replace the perturbation-based samples. 
Different from perturbation-based method that randomly discards the information of the graph, the proposed generation module could maintain the integrity of the graph.
Our generation module, by assigning the learned attentional weights to the neighborhood nodes, can adaptively exploit the intrinsic geometric structure of the graph and generate more effective contrastive samples.
Besides, since the similarity between adjacent nodes in the graph is an inherent attribute, there is a strong correlation between the central node and its neighborhood.
Therefore, the generated subgraphs by neighborhood interpolation are inherently similar to the original subgraphs.
And it is reasonable and effective to treat the generated subgraph as positive sample.

%-------------------------------------------------------------------------
\subsection{OT Distance Based Contrastive Learning}
Most graph contrastive learning methods use node pairs or node-subgraph pairs or subgraph pairs as contrastive samples. 
Particularly, the features of subgraphs can be extracted with the readout function. 
Thus, these methods mainly employ the vector-wise similarity metrics to calculate the similarity between these samples. 
However, the vector inner product based similarity metrics cannot fully exploit the local structures of the graph to characterize the graph difference well.
Instead of using the vector-wise similarity metrics, in our method, we introduce the optimal transport distance (i.e., Wasserstein distance and Gromov-Wasserstein distance) as the similarity metric for contrastive subgraphs. 
Therefore, we can accurately characterize the geometric difference between the subgraphs.

\textbf{Wasserstein distance (WD).} WD is commonly used for matching two discrete distributions (e.g., two sets of node embeddings) \cite{chen2020graph}. 
It can represent the cost of converting one subgraph to another by counting the difference between all node pairs in the two subgraphs. 
In our settings, WD is employed to measure the similarity between the nodes of the subgraphs.
The WD for similarity calculation between subgraphs can be described as follows.

Let $\bm{u}$ and $\bm{v}$ represent discrete distributions of two subgraphs, where $\bm{u}=\{ u_{1}, u_{2},..., u_{n} \}$ and $\bm{v}=\{v_{1}, v_{2},..., v_{m}\}$, $\sum^{n}_{i=1}u_{i}\! =\! \sum^{m}_{j=1}v_{j}\! =\! 1$, $n$ and $m$ are the number of the subgraph nodes, respectively. The WD between the two discrete distributions $\bm{u}$ and $\bm{v}$ can be defined as: \begin{equation}\label{key}
	D_{w}(\bm{u}, \bm{v}) = \min_{\bm{T}\in \pi(\bm{u}, \bm{v})}\sum^{n}_{i=1}\sum^{m}_{j=1}\bm{T}_{ij}c(\bm{h}_{1i}, \bm{h}_{2j})
\end{equation}
where $\pi(\bm{u}, \bm{v})$ represents all the joint distributions between two subgraphs nodes. $c(\bm{h}_{1i}, \bm{h}_{2j})= exp(-\varphi(\bm{h}_{1i}, \bm{h}_{2j}) / \tau) $ denotes the transport cost between node $i$ in subgraph $1$ and node $j$ in subgraph $2$,
$\bm{h}_{1i}$ and $\bm{h}_{2j}$ represent the node features, $\tau$ is a temperature parameter, and $\varphi(.,.)$ denotes the cosine similarity between the node features. 
The matrix $\bm{T}$ represents the transport plan, where $\bm{T}_{ij}$ denotes the amount of mass shifted from $u_{i}$ to $v_{j}$. 
And $\bm{T}$ can be achieved by applying the Sinkhorn algorithm \cite{cuturi2013sinkhorn,peyre2019computational} with an entropic regularizer \cite{benamou2015iterative}.
\begin{equation}\label{key}
	\min_{\bm{T}\in \pi(\bm{u}, \bm{v})}\sum^{n}_{i=1}\sum^{m}_{j=1}\bm{T}_{ij}c(\bm{h}_{1i}, \bm{h}_{2j}) + \beta H(\bm{T})
\end{equation}
where $H(\bm{T})= \sum_{i,j}\bm{T}_{ij} log\bm{T}_{ij}$, and $\beta$ is the hyperparameter controlling the importance of the entropy term.

\textbf{WD-based contrastive loss.} Based on the Wasserstein distance, we define the loss as:
\begin{equation}\label{key}
	\begin{aligned}
		\mathcal{L}_{1} \!=\!  \frac{-1}{N(M \!+\! 1)} \sum^{N}_{i=1}\lbrack  log(exp(- D_{w}(\bm{s}_{i}, \bm{s}_{p})  / \tau )) \!+\! 
		\sum^{M}_{j=1} log(1 \!- \!exp( \!- \! D_{w}(\bm{s}_{i},\bm{s}_{nj})/ \tau)   )    \rbrack
	\end{aligned}
\end{equation}
where $N$ is the number of sampled subgraphs, $M$ is the number of negative samples of each subgraph, and $\tau$ is a temperature parameter. %, we set $\tau=0.01$. 
$(s_{i}, s_{p})$ denotes the positive sample pair, $(s_{i}, s_{nj})$ denotes the negative sample pair. 
To speed up the calculation efficiency, we only randomly select two negative samples, i.e., $M\! =\! 2$, one from the sampled subgraphs and the other from the generated subgraphs.

Compared with readout function-based manner, WD can exploit similar information among all nodes and distinguish the contrastive samples more effectively.
Therefore, with the WD-based contrastive loss, we can maximize the similarity between the nodes across the positive subgraphs and minimize the similarity between the nodes across the negative subgraphs.

\textbf{Gromov-Wasserstein distance (GWD)}. 
Unlike WD, which can directly calculate the distance of node pairs between two subgraphs, GWD \cite{chowdhury2019gromov,peyre2016gromov} can be used when we can only get the distances between pairs of nodes within each subgraph. 
GWD can be used to calculate the distance between node pairs within the subgraph, as well as to measure the differences in these distances across the subgraphs. 
That is to say, GWD can measure the distances between node pairs within each subgraph compare to those in the counterpart subgraph. 
Therefore, GWD can be used to capture the similarity between the edges of the subgraphs.
The GWD for similarity calculation between subgraphs can be described as:

Let $\bm{u}$ and $\bm{v}$ represent discrete distributions of two subgraphs, where $\bm{u}=\{u_{1}, u_{2},..., u_{n}\}$ and $\bm{v}=\{v_{1}, v_{2},..., v_{m}\}$, $\sum^{n}_{i=1}u_{i}\! =\! \sum^{m}_{j=1}v_{j}\! =\! 1$, $n$ and $m$ is the number of subgraph nodes.
The GWD between the two discrete distributions $\bm{u}$, $\bm{v}$ can be defined as:
\begin{equation}\label{key}
	D_{gw}(\bm{u}, \bm{v}) = \min_{\bm{T}\in \pi(\bm{u}, \bm{v})}\sum_{i,i^{'}\!,j,j^{'}} \bm{T}_{ij}\bm{T}_{i^{'}\!j^{'}}\hat{c}(\bm{h}_{1i}, \bm{h}_{2j}, \bm{h}_{1i^{'}}, \bm{h}_{2j^{'}})
\end{equation}
where $\pi(\bm{u}, \bm{v})$ denotes all the joint distributions, the matrix $\bm{T}$ represents the transport plan between two subgraphs, $\bm{T}_{ij}$ denotes the amount of mass shifted from $u_{i}$ to $v_{j}$.
$\hat{c}(\bm{h}_{1i}, \bm{h}_{2j}, \bm{h}_{1i^{'}}, \bm{h}_{2j^{'}})= \Vert c(\bm{h}_{1i}, \bm{h}_{1i^{'}}) - c(\bm{h}_{2j}, \bm{h}_{2j^{'}}) \Vert_2$ is the cost function to measure  the edge difference between two subgraphs.
$c(.,.)$ represents the distance between nodes within the subgraph.

Given the adjacent matrix of the sampled subgraph, for the nodes $s_1$ and $s_2$ of sampled subgraph $s$, the distance $c(\bm{h}_{s_1}, \bm{h}_{s_2})$ can be defined as:
\begin{equation}\label{key}
	c(\bm{h}_{s_1}, \bm{h}_{s_2}) = exp(- \bm{A}_{s}(s_1, s_2) / \tau )
\end{equation}
where $\bm{A}_{s}(s_1, s_2)$ represents the connection relationship between node $s_1$ and node $s_2$ of the sampled subgraph, $\tau$ is a temperature parameter.

The distance between the generated subgraph nodes can be defined as:
\begin{equation}\label{key}
	\begin{aligned}
		c(\bm{\hat{h}}_{s_1}, \bm{\hat{h}}_{s_2})   
		= exp(- \bm{\hat{A}}_{s}(s_1,s_2) / \tau ) \\
	\end{aligned}
\end{equation}
where $\bm{\hat{A}}_{s}(s_1,s_2) = \varphi(\bm{\hat{h}}_{s_1}, \bm{\hat{h}}_{s_2})$ represents the connection relationship between the node $s_1$ and node $s_2$ of generated subgraph, $\varphi(.,.)$ represents the consine similarity, $\tau$ is a temperature parameter. 

\textbf{GWD-based contrastive loss.}
Based on the Gromov-Wasserstein distance, we define the loss as:
\begin{equation}\label{key}
	\begin{aligned}
		\mathcal{L}_{2} = \frac{-1}{N \!( \! M \!+\!1 \!)\!} \sum^{N}_{i=1} \lbrack  log(exp(\!- D \!_{gw}(\bm{s}_{i},   \bm{s}_{p}) / \tau)) \!+ \!
		\sum^{M}_{j=1} log(1 \!-\! exp( \! -D \!_{gw}(\bm{s}_{i}, \bm{s}_{nj})/ \tau)   )    \rbrack
	\end{aligned}
\end{equation}
where $N$ is the number of sampled subgraphs, $M$ is the number of negative samples of each subgraph and we also set $M=2$, $\tau$ is a temperature parameter.
$(s_{i}, s_{p})$ and $(s_{i}, s_{nj})$ denote positive and negative sample pairs.
The GWD-based contrastive loss can maximize the similarity between the edges of the positive subgraphs and minimize the similarity between the edges of the negative subgraphs so that the geometry difference between the subgraphs can be captured.

Finally, we obtain the final loss function $\mathcal{L}$, which is defined as follows:
\begin{equation}\label{key_L}
	\mathcal{L} = \lambda \mathcal{L}_{1} + (1-\lambda)\mathcal{L}_{2}
\end{equation}
where $\lambda$ is the hyper-parameter for controlling the importance of different loss functions. Here, we set $\lambda=0.5$.
To the best of our knowledge, we are the first to introduce OT into subgraph-based graph contrastive learning.
We use WD to exploit contrastive information based on the subgraph node features, and GWD to exploit contrastive information based on the edges of the subgraph.
Therefore, the OT-based contrastive loss can better guide the training of the encoder.

%------------------------------------------------------------------------
\section{Experiment}

In this section, extensive experiments are conducted to evaluate the performance of our method in a self-supervised manner on the transductive and inductive node classification tasks. And we compare our method with other baselines, including unsupervised and supervised methods. Besides, we conduct ablation studies to verify the effectiveness of our proposed method.

\subsection{Datasets}
In order to evaluate the effectiveness of our method, we conduct experiments on five benchmark datasets of three real-world networks. 
Following \cite{hamilton2017inductive,kipf2016semi}, three tasks are performed: 
(1) classifying the topics of the documents on the citation network datasets of Cora, Citeseer and Pubmed \cite{sen2008collective}; 
(2) classifying protein roles of protein-protein interaction (PPI) networks \cite{zitnik2017predicting}, and generalizing to unseen networks; 
(3) predicting the community structure of the social network on Reddit posts \cite{zeng2019graphsaint}. 
More detailed descriptions can be found in the supplementary materials.

\subsection{Implementation details}
Due to different attributes of datasets, we employ distinct encoders for three experimental settings, i.e., transductive learning for the small graph, inductive learning for the large graph and multiple graphs.
For more specific information of encoders may be found in \cite{velivckovic2018deep}, or please refer to the supplementary materials.

All experiments are implemented using PyTorch \cite{chollet2017deep} and the geometric deep learning extension library \cite{fey2019fast}. The experiments are conducted
on  a single TITAN RTX GPU. Our method is used to learn node representations in a self-supervised manner, followed by evaluating the learned representations with the node classification task. This is performed by training and testing a simple linear (logistic regression) classifier in the downstream tasks using the learned representations. 
We train the model by minimizing the loss function provided in Eq. \hyperref[key_L]{(12)}. 
And we use Adam optimizer \cite{kingma2014adam} with an initial learning rate of 0.0001 (especially, 0.0005 on Pubmed). 
The dimension of node representations is 1024 (256 with 4 heads for PPI). 
In order to avoid the excessive calculation, in every epoch, we randomly sample some subgraphs to calculate the OT distance for the loss calculation.
Besides, parameter $\bm{T}$ is shared by WD and GWD.
The detailed parameter settings can be found in the supplementary materials.

\subsection{Results}
\textbf{Classification results.}
We choose four state-of-the-art graph contrastive learning methods to evaluate graph embeddings, DGI \cite{velivckovic2018deep}, GMI \cite{peng2020graph}, GraphCL \cite{hafidi2020graphcl} and Subg-Con \cite{jiao2020sub}.
And two traditional unsupervised methods, DeepWalk \cite{perozzi2014deepwalk} and the unsupervised variant of GraphSAGE \cite{hamilton2017inductive} are also compared with our method. Specially, we also provide results for training the classifier on the raw input features. Besides, we report the results of three supervised graph neural networks, GCN \cite{kipf2016semi}, GAT \cite{velivckovic2017graph} and GraphSAGE \cite{hamilton2017inductive}.
For the node classification task, we employ mean classification accuracy to evaluate the performance on Cora, Citeseer, and Pubmed datasets, while the micro-averaged F1 score for the Reddit and PPI datasets.

The evaluation results on the five datasets are listed in Table. \hyperref[results]{1}. 
The results demonstrate that our method has achieved good performance across all five datasets. 
As can be seen in Table. \hyperref[results]{1}, our method successfully outperforms all the competing graph contrastive learning approaches, which implies the potential of our proposed method for the node classification task.
Compared with the traditional subgraph perturbation based GraphCL, our method has the gain of at least $1\%$ improvement on all data sets, even $3.2\%$ on PPI. 
This indicates that our subgraph generation module can effectively capture the intrinsic local structures of the graph.
We further observe that the performance of our method is better than other vector inner product based methods, which verifies that our OT-based similarity metric can effectively characterize the graph difference by exploiting local structures of the graph.
From this table, one can also see that although the labels of the nodes are used in the supervised graph representation learning methods, the proposed graph contrastive learning method can still outperform most of the supervised methods.

\begin{table*}[t]
	\caption{Performance comparison  of node classification  with different methods on the transductive and inductive tasks. The third column illustrates the data used by each algorithm in the training phase, where X, A, and Y denote features, adjacency matrix, and labels, respectively. For simple expression, we abbreviate our method as GSC.}
	\label{results}
	\centering
	\resizebox{0.98\textwidth}{!}{
		\begin{tabular}{cccccccc}
			\toprule  
			&\multirow{2}{*}{Algorithm}  &\multirow{2}{*}{Data} &\multicolumn{3}{c}{Transductive}  &\multicolumn{2}{c}{Inductive} \\ \cmidrule(r){4-6}  \cmidrule(r){7-8}
			& & &Cora &Citeseer &Pubmed &PPI &Reddit \\
			
			\midrule      
			\multirow{3}{*}{Supervised}&GCN \cite{kipf2016semi}&$\bm{X,A,Y}$ &81.4$\pm$0.6 &70.3$\pm$0.7 &76.8$\pm$0.6 &51.5$\pm$0.6 &93.3$\pm$0.1\\
			&GAT \cite{velivckovic2017graph}&$\bm{X,A,Y}$ &83.0$\pm$0.7 &72.5$\pm$0.7 &79.0$\pm$0.3 &\textbf{97.3$\pm$0.2} &-\\
			&GraphSAGE \cite{hamilton2017inductive} &$\bm{X,A,Y}$ &79.2$\pm$1.5 &71.2$\pm$0.5 &73.1$\pm$1.4 &51.3$\pm$3.2 &92.1$\pm$1.1\\
			
			\midrule     
			\multirow{7}{*}{Unsupervised} &Raw features &$\bm{X}$ &56.6$\pm$0.4 &57.8$\pm$0.2 &69.1$\pm$0.2 &42.5$\pm$0.3 &58.5$\pm$0.1\\ 
			&DeepWalk \cite{perozzi2014deepwalk} &$\bm{A}$ &67.2 &43.2 &65.3 &52.9 &32.4\\
			&GraphSAGE \cite{hamilton2017inductive} &$\bm{X,A}$ &75.2$\pm$1.5 &59.4$\pm$0.9 &70.1$\pm$1.4 &46.5$\pm$0.7 &90.8$\pm$1.1 \\		
			&DGI \cite{velivckovic2018deep}&$\bm{X,A}$&82.3$\pm$0.6 &71.8$\pm$0.7 &76.8$\pm$0.6 &63.8$\pm$0.2 &94.0$\pm$0.1 \\
			&GMI \cite{peng2020graph}&$\bm{X,A}$&83.0$\pm$0.3 &73.0$\pm$0.3 &79.9$\pm$0.2 &65.0$\pm$0.0 &95.0$\pm$0.0 \\
			&GraphCL \cite{hafidi2020graphcl}&$\bm{X,A}$&83.6$\pm$0.5 &72.5$\pm$0.7 &79.8$\pm$0.5 &65.9$\pm$0.6 &95.1$\pm$0.1 \\
			&Subg-Con \cite{jiao2020sub}&$\bm{X,A}$ &83.5$\pm$0.5 &73.2$\pm$0.2 &81.0$\pm$0.1 &66.9$\pm$0.2 &95.2$\pm$0.0 \\			
			
			\cmidrule{2-8} 
			&GSC (ours) &$\bm{X,A}$ &\textbf{84.6$\pm$0.1} &\textbf{73.7$\pm$0.1} &\textbf{82.1$\pm$0.2} &\textbf{69.1$\pm$0.3} &\textbf{95.3$\pm$0.1} \\
			\bottomrule   
	\end{tabular}}
\end{table*}

We also conduct node classification experiments with few contrastive training samples to evaluate our method. 
When using $70\%$, $50\%$, $30\%$ and $10\%$ contrastive samples on the Cora dataset, the performance of Subg-Con is $83.0\%$, $82.3\%$, $81.2\%$ and $79.6\%$, while  the performance of GraphCL is $83.2\%$, $82.5\%$, $81.6\%$ and $80.1\%$, respectively.
Nonetheless, by generating multiple positive samples for each original subgraph, the performance of our method is $84.6\%$, $84.1\%$, $83.6\%$ and $82.9\%$, respectively. Particularly, in the case of $10\%$ contrastive samples, our method can obtain the gain of $3.3\%$.  
%Since Subg-Con can only construct positive samples by pooling the top-k important neighbors, the performance is limited. 
Since Subg-Con can only construct one positive sample by pooling the top-k important neighbors for each subgraph, the performance is limited in the cases of few contrastive training samples. 
Different from Subg-Con, based on neighborhood interpolation, our subgraph generation module can generate multiple positive samples for each samples, which can effectively handle the situation of insufficient contrastive samples. 
Although GraphCL can also construct multiple positive samples, the perturbation-based data augmentation manner may change the original attributes of the graph.
Compared with GraphCL, our neighborhood interpolation based generation module can effectively preserve the local structures of the graph.
For more results, please refer to supplementary materials.

\begin{figure}[t]
	\centering
	\subfigure[Raw features.]{
		\includegraphics[width=2.5cm]{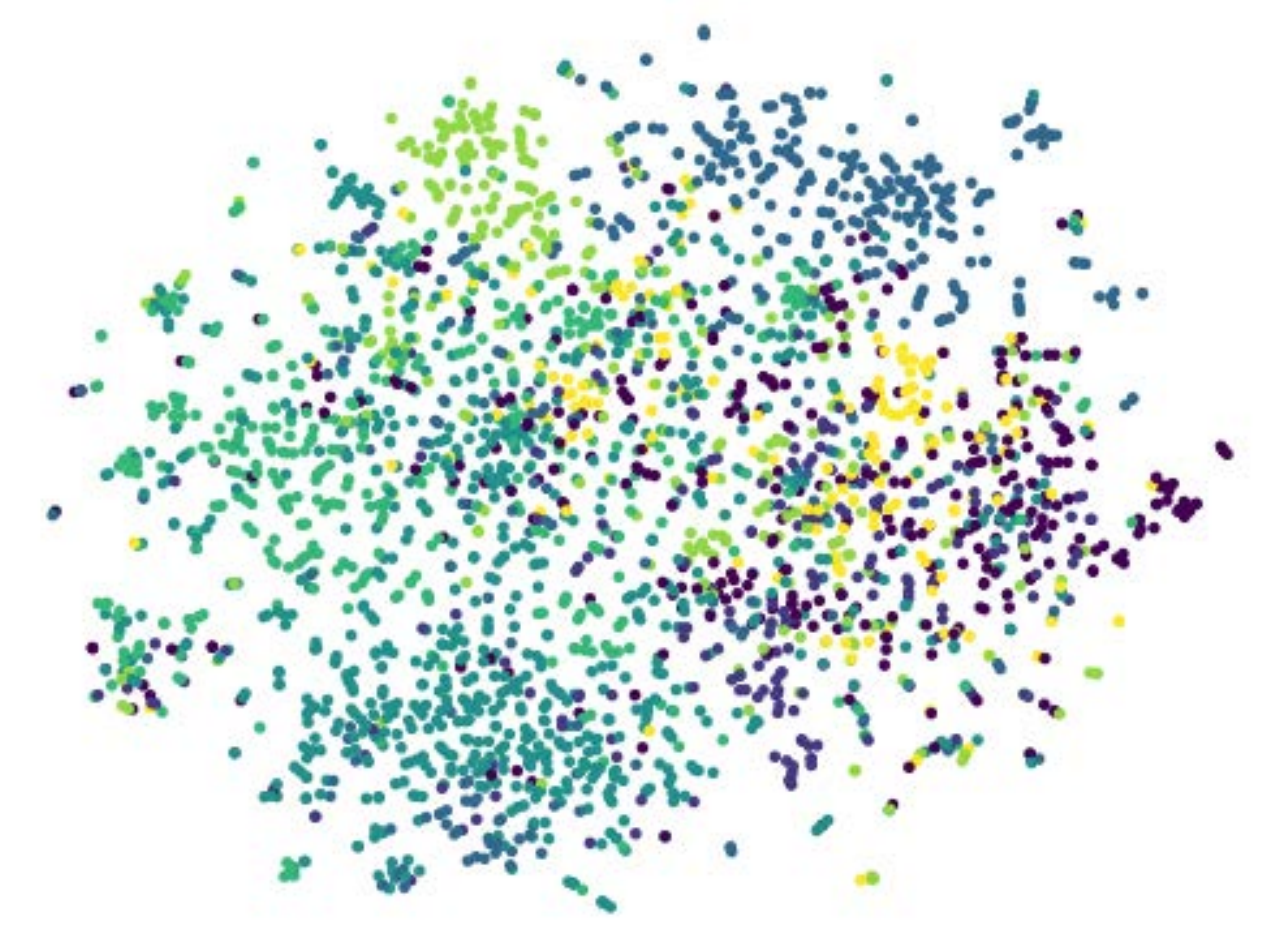}
	}
	\quad
	\subfigure[DGI.]{
		\includegraphics[width=2.5cm]{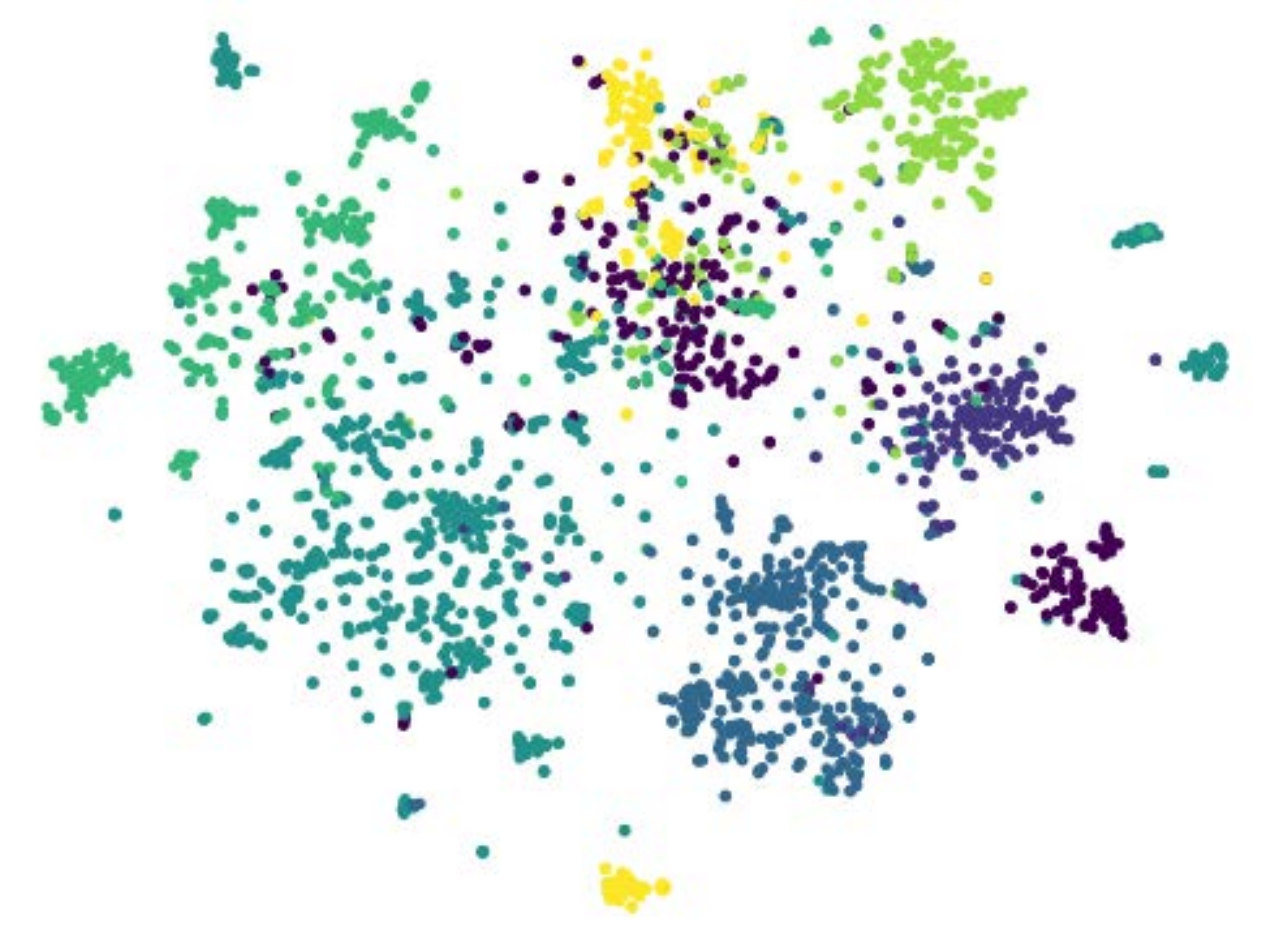}
	}
	\quad
	\subfigure[Subg-Con.]{
		\includegraphics[width=2.5cm]{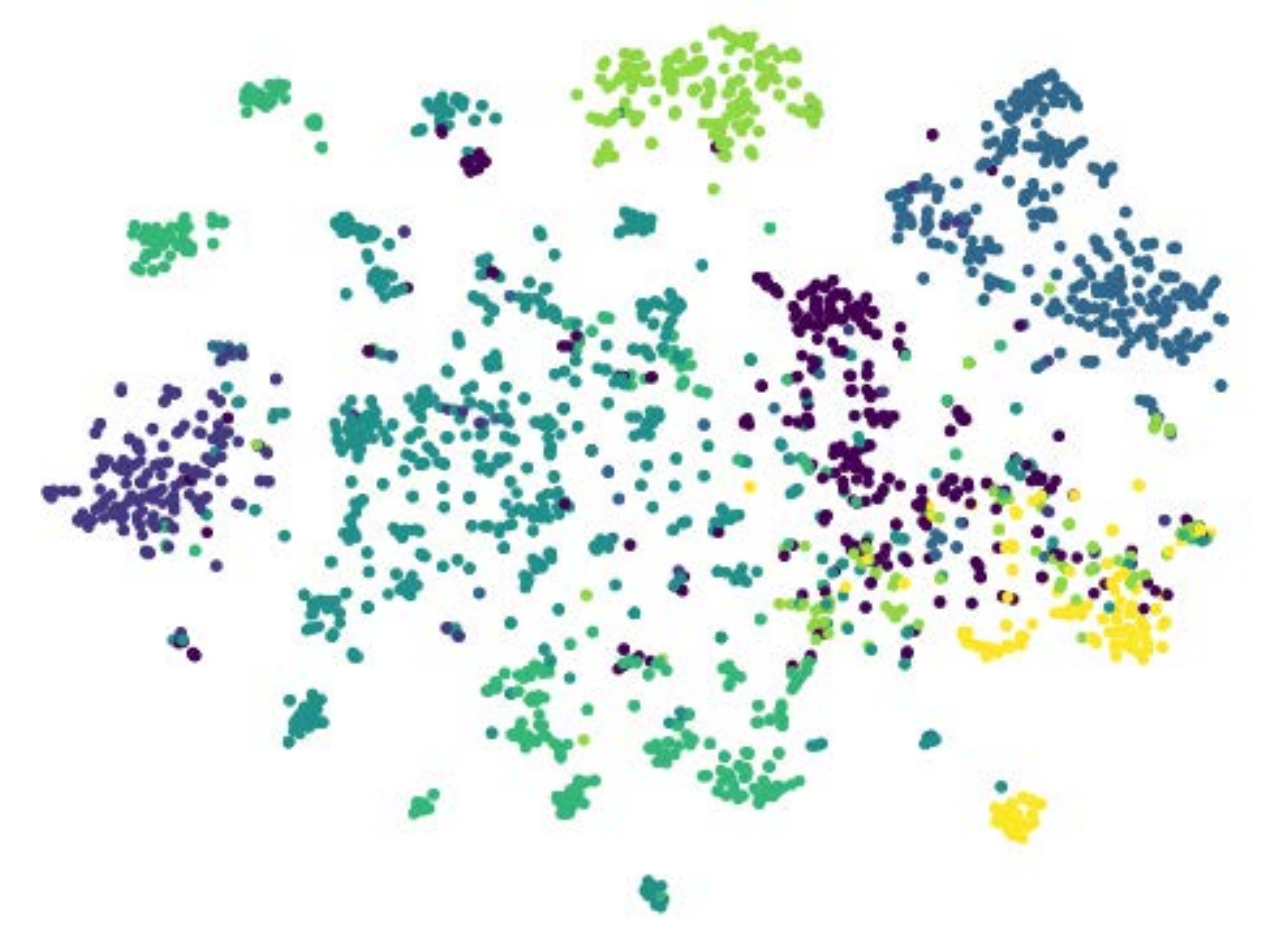}
	}
	\quad
	\subfigure[GSC (ours).]{
		\includegraphics[width=2.5cm]{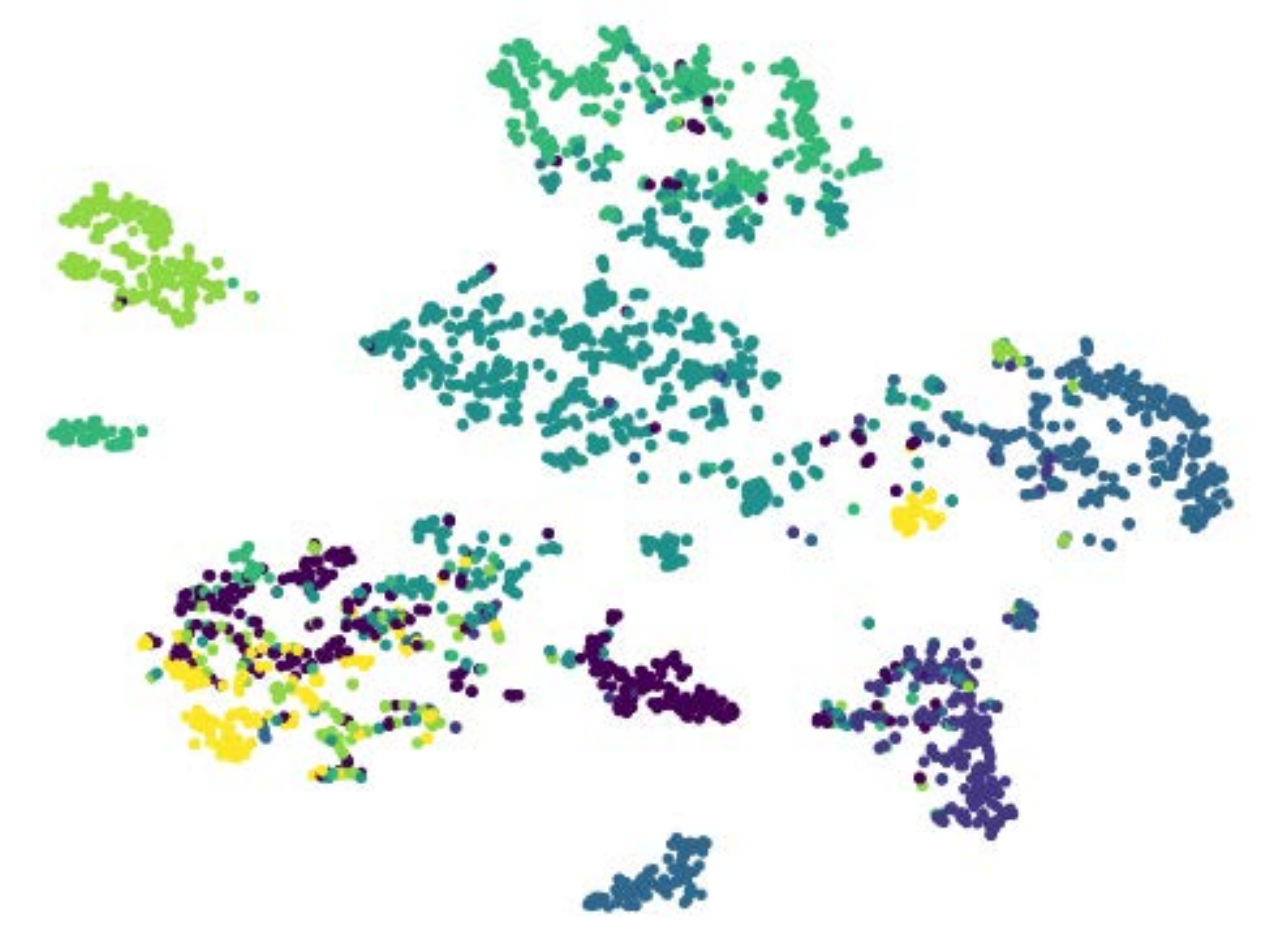}
	}
	\caption{Visualization of t-SNE embeddings of raw features, DGI, Sub-Con, and GSC (ours) on Cora dataset.} 
	\label{fig:Cora}
\end{figure}

\textbf{Visual results.}
As shown in Fig. \hyperref[fig:Cora]{3}, we also visualize the raw features and learned embeddings of the graphs with the t-SNE \cite{van2008visualizing} plot for different graph contrastive learning methods, including DGI, Subg-Con and GSC (ours). From the visualization results in Fig. \hyperref[fig:Cora]{3}, one can see that the embeddings generated by GSC can exhibit closer clusters than the other three methods.
This means that our graph contrastive learning can obtain more discriminative features.

\subsection{Ablation studies}
\textbf{Effectiveness of generation module and OT distance.}
To further verify the effectiveness of different modules, we conduct three sets of comparative experiments on Cora, Citeseer and, Pubmed datasets.  
To verify the effectiveness of the OT distance based contrastive loss, we use the readout function to obtain the feature vectors of the subgraphs and calculate the vector-wise similarity.
As can be seen from Table. \hyperref[Ablation11]{2}, in comparison with Readout, the use of the OT distance can improve the classification performance.
This demonstrates that the OT distance can effectively capture local structure information of the graph and distinguish different subgraphs. 

To demonstrate the effectiveness of the generation module, we use the traditional perturbation on the subgraph to replace the generation module and calculate the similarity between subgraphs with the OT distance. 
From Table. \hyperref[Ablation11]{2}, we can see that the classification performance of Generation + OT outperforms that of Perturbation + OT. 
This can verify that our generation module can effectively capture the intrinsic local structures of the graph. The generated samples can improve the performance of graph contrastive learning. 
\\
\begin{minipage}{\textwidth}
	\begin{minipage}[t]{0.49\textwidth}
		\makeatletter\def\@captype{table}
		\caption{Ablation studies on different modules}
		\label{Ablation11}
		\resizebox{0.97\textwidth}{!}{
			\begin{tabular}{ccccc}
				\toprule  
				&  &Cora &Citeseer &Pubmed  \\
				\midrule      
				&Gene + Readout &83.6 &72 &78.5  \\
				&Perturbation + OT &84.1  &72.3 &80.4 \\
				&Gene + OT (ours) &\textbf{84.6} &\textbf{73.7} &\textbf{82.1} \\
				\bottomrule   
		\end{tabular}}
	\end{minipage}
	\begin{minipage}[t]{0.46\textwidth}
		\makeatletter\def\@captype{table}
		\caption{Ablation studies on different sampling methods}
		\label{Ablation12}
		\resizebox{0.95\textwidth}{!}{
			\begin{tabular}{ccccc}
				\toprule  
				&  &Cora &Citeseer &Pubmed  \\
				\midrule     
				&Importance Score &83.5 &71.68 &80.6  \\
				&Random Walk &84.2 &72.4 &81.2  \\
				&BFS (ours) &\textbf{84.6} &\textbf{73.7} &\textbf{82.1}  \\
				\bottomrule   
		\end{tabular}}
	\end{minipage}
\end{minipage}\\

\textbf{Effectiveness of different sampling methods.}
We compare different subgraph sampling methods (i.e., importance score \cite{jiao2020sub} and random walk \cite{you2020graph} ) and list the experiment results in Table \hyperref[Ablation12]{3}.
To make the comparisons fair, we sample subgraphs of the same size.
As can be seen from Table. \hyperref[Ablation12]{3}, BFS-based method can achieve the best performance compared with other sampling methods.
This can verify that BFS-based sampled subgraphs can better cover local information and can be more beneficial to distinguish between subgraphs.

\begin{figure}[t]
	\centering
	\label{fig:aba} 
	\subfigure[Different subgraph sizes]{
		\includegraphics[width=5.2cm]{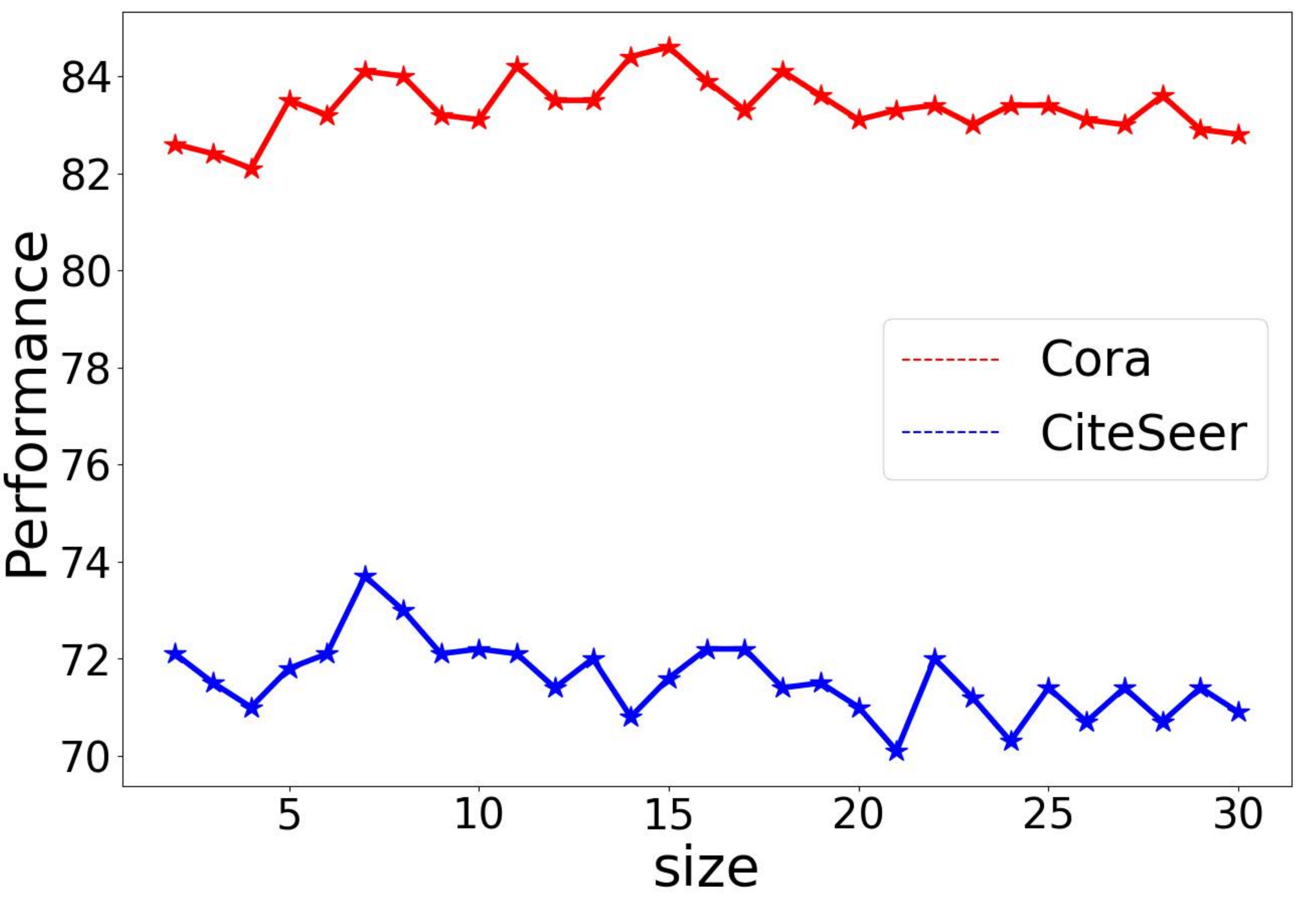}
	}
	\quad
	\subfigure[Different $\lambda$]{
		\includegraphics[width=5.2cm]{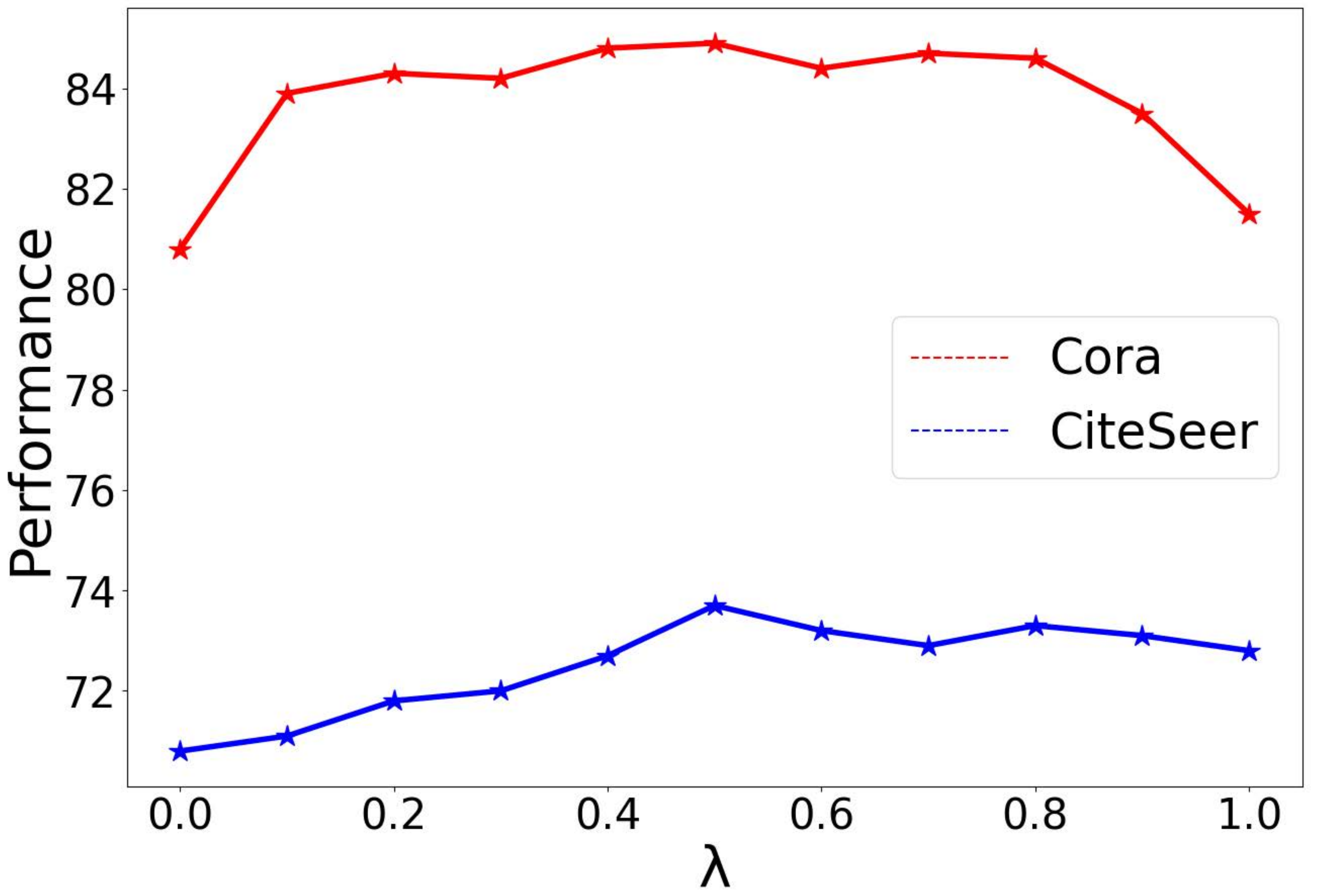} 
	}
	\caption{Ablation studies of different subgraph sizes and parameter $\lambda$.}
\end{figure}

\textbf{Influence of the subgraph size and parameter $\lambda$.}
To study the influence of the subgraph size in our method, conduct experiments on the Cora and Citeseer datasets by varying the numbers of subgraph nodes from 2 to 30.
As can be seen from Fig. \hyperref[fig:aba]{4} (a), the classification performance is slightly fluctuated with different subgraph sizes.
Besides, we vary the parameter $\lambda$ from 0 to 1 to study the effects on the final classification accuracy.
Different $\lambda$ can control different weights of the loss terms. 
As shown in Fig. \hyperref[fig:aba]{4} (b), both node and edge features have effects on the final performance and the classification performance can be kept relatively stable in [0.4, 0.8]. 
The best performance can be achieved when $\lambda$ is set to 0.5, where the contributions of node and edge features reach a balance.

%-------------------------------------------------------------------------
\section{Conclusion}
In this paper, we proposed a novel subgraph generation method for graph contrastive learning.
Based on the neighborhood interpolation, we developed the subgraph generation module, which can adaptively generate the contrastive subgraphs.
Furthermore, we employed two types of optimal transport distances (i.e., Wasserstein distance and Gromov-Wasserstein distance) to calculate the similarity between subgraphs. 
In particular, generated subgraphs and the OT distance based similarity metric can effectively capture the intrinsic local structures of the graph to characterize the graph difference well.
By conducting extensive experiments on multiple benchmark datasets, we demonstrate that our proposed graph contrastive learning method can yield better performance in comparison with the supervised and unsupervised graph representation learning methods.

%\section*{Acknowledgments}
%The authors would like to thank reviewers for their detailed comments and instructive suggestions.
%This work was supported by the National Science Fund of China (Grant Nos. 61876084, 61876083, 62176124).

\clearpage
%%%%%%%%% REFERENCES
\bibliographystyle{splncs04}
\bibliography{egbib}

\begin{thebibliography}{10}
\providecommand{\url}[1]{\texttt{#1}}
\providecommand{\urlprefix}{URL }
\providecommand{\doi}[1]{https://doi.org/#1}

\bibitem{benamou2015iterative}
Benamou, J.D., Carlier, G., Cuturi, M., Nenna, L., Peyr{\'e}, G.: Iterative
  bregman projections for regularized transportation problems. SIAM Journal on
  Scientific Computing  \textbf{37}(2),  A1111--A1138 (2015)

\bibitem{chen2020graph}
Chen, L., Gan, Z., Cheng, Y., Li, L., Carin, L., Liu, J.: Graph optimal
  transport for cross-domain alignment. In: International Conference on Machine
  Learning. pp. 1542--1553. PMLR (2020)

\bibitem{chollet2017deep}
Chollet, F.: Deep learning with Python. Simon and Schuster (2017)

\bibitem{chowdhury2019gromov}
Chowdhury, S., M{\'e}moli, F.: The gromov--wasserstein distance between
  networks and stable network invariants. Information and Inference: A Journal
  of the IMA  \textbf{8}(4),  757--787 (2019)

\bibitem{Chu2021CuCoGR}
Chu, G., Wang, X., Shi, C., Jiang, X.: Cuco: Graph representation with
  curriculum contrastive learning. In: IJCAI (2021)

\bibitem{cuturi2013sinkhorn}
Cuturi, M.: Sinkhorn distances: Lightspeed computation of optimal transport.
  Advances in neural information processing systems  \textbf{26},  2292--2300
  (2013)

\bibitem{faerman2019graph}
Faerman, E., Voggenreiter, O., Borutta, F., Emrich, T., Berrendorf, M.,
  Schubert, M.: Graph alignment networks with node matching scores. Proceedings
  of Advances in Neural Information Processing Systems (NIPS)  (2019)

\bibitem{fey2019fast}
Fey, M., Lenssen, J.E.: Fast graph representation learning with pytorch
  geometric. arXiv preprint arXiv:1903.02428  (2019)

\bibitem{hafidi2020graphcl}
Hafidi, H., Ghogho, M., Ciblat, P., Swami, A.: Graphcl: Contrastive
  self-supervised learning of graph representations. arXiv preprint
  arXiv:2007.08025  (2020)

\bibitem{hamilton2017inductive}
Hamilton, W.L., Ying, R., Leskovec, J.: Inductive representation learning on
  large graphs. In: Proceedings of the 31st International Conference on Neural
  Information Processing Systems. pp. 1025--1035 (2017)

\bibitem{hamilton2017representation}
Hamilton, W.L., Ying, R., Leskovec, J.: Representation learning on graphs:
  Methods and applications. arXiv preprint arXiv:1709.05584  (2017)

\bibitem{hassani2020contrastive}
Hassani, K., Khasahmadi, A.H.: Contrastive multi-view representation learning
  on graphs. In: International Conference on Machine Learning. pp. 4116--4126.
  PMLR (2020)

\bibitem{hjelm2018learning}
Hjelm, R.D., Fedorov, A., Lavoie-Marchildon, S., Grewal, K., Bachman, P.,
  Trischler, A., Bengio, Y.: Learning deep representations by mutual
  information estimation and maximization. arXiv preprint arXiv:1808.06670
  (2018)

\bibitem{jiao2020sub}
Jiao, Y., Xiong, Y., Zhang, J., Zhang, Y., Zhang, T., Zhu, Y.: Sub-graph
  contrast for scalable self-supervised graph representation learning. arXiv
  preprint arXiv:2009.10273  (2020)

\bibitem{jin2020self}
Jin, W., Derr, T., Liu, H., Wang, Y., Wang, S., Liu, Z., Tang, J.:
  Self-supervised learning on graphs: Deep insights and new direction. arXiv
  preprint arXiv:2006.10141  (2020)

\bibitem{kingma2014adam}
Kingma, D.P., Ba, J.: Adam: A method for stochastic optimization. arXiv
  preprint arXiv:1412.6980  (2014)

\bibitem{kipf2016semi}
Kipf, T.N., Welling, M.: Semi-supervised classification with graph
  convolutional networks. arXiv preprint arXiv:1609.02907  (2016)

\bibitem{kipf2016variational}
Kipf, T.N., Welling, M.: Variational graph auto-encoders. arXiv preprint
  arXiv:1611.07308  (2016)

\bibitem{lee2019self}
Lee, J., Lee, I., Kang, J.: Self-attention graph pooling. In: International
  Conference on Machine Learning. pp. 3734--3743. PMLR (2019)

\bibitem{lee2022augmentation}
Lee, N., Lee, J., Park, C.: Augmentation-free self-supervised learning on
  graphs. In: Proceedings of the AAAI Conference on Artificial Intelligence.
  vol.~36, pp. 7372--7380 (2022)

\bibitem{van2008visualizing}
Van~der Maaten, L., Hinton, G.: Visualizing data using t-sne. Journal of
  machine learning research  \textbf{9}(11) (2008)

\bibitem{manessi2021graph}
Manessi, F., Rozza, A.: Graph-based neural network models with multiple
  self-supervised auxiliary tasks. Pattern Recognition Letters  \textbf{148},
  15--21 (2021)

\bibitem{peng2020graph}
Peng, Z., Huang, W., Luo, M., Zheng, Q., Rong, Y., Xu, T., Huang, J.: Graph
  representation learning via graphical mutual information maximization. In:
  Proceedings of The Web Conference 2020. pp. 259--270 (2020)

\bibitem{perozzi2014deepwalk}
Perozzi, B., Al-Rfou, R., Skiena, S.: Deepwalk: Online learning of social
  representations. In: Proceedings of the 20th ACM SIGKDD international
  conference on Knowledge discovery and data mining. pp. 701--710 (2014)

\bibitem{peyre2016gromov}
Peyr{\'e}, G., Cuturi, M., Solomon, J.: Gromov-wasserstein averaging of kernel
  and distance matrices. In: International Conference on Machine Learning. pp.
  2664--2672. PMLR (2016)

\bibitem{peyre2019computational}
Peyr{\'e}, G., Cuturi, M., et~al.: Computational optimal transport: With
  applications to data science. Foundations and Trends{\textregistered} in
  Machine Learning  \textbf{11}(5-6),  355--607 (2019)

\bibitem{qiu2020gcc}
Qiu, J., Chen, Q., Dong, Y., Zhang, J., Yang, H., Ding, M., Wang, K., Tang, J.:
  Gcc: Graph contrastive coding for graph neural network pre-training. In:
  Proceedings of the 26th ACM SIGKDD International Conference on Knowledge
  Discovery \& Data Mining. pp. 1150--1160 (2020)

\bibitem{sen2008collective}
Sen, P., Namata, G., Bilgic, M., Getoor, L., Galligher, B., Eliassi-Rad, T.:
  Collective classification in network data. AI magazine  \textbf{29}(3),
  93--93 (2008)

\bibitem{sun2019infograph}
Sun, F.Y., Hoffmann, J., Verma, V., Tang, J.: Infograph: Unsupervised and
  semi-supervised graph-level representation learning via mutual information
  maximization. arXiv preprint arXiv:1908.01000  (2019)

\bibitem{suresh2021adversarial}
Suresh, S., Li, P., Hao, C., Neville, J.: Adversarial graph augmentation to
  improve graph contrastive learning. arXiv preprint arXiv:2106.05819  (2021)

\bibitem{velivckovic2017graph}
Veli{\v{c}}kovi{\'c}, P., Cucurull, G., Casanova, A., Romero, A., Lio, P.,
  Bengio, Y.: Graph attention networks. arXiv preprint arXiv:1710.10903  (2017)

\bibitem{velivckovic2018deep}
Veli{\v{c}}kovi{\'c}, P., Fedus, W., Hamilton, W.L., Li{\`o}, P., Bengio, Y.,
  Hjelm, R.D.: Deep graph infomax. arXiv preprint arXiv:1809.10341  (2018)

\bibitem{wang2022augmentation}
Wang, H., Zhang, J., Zhu, Q., Huang, W.: Augmentation-free graph contrastive
  learning. arXiv preprint arXiv:2204.04874  (2022)

\bibitem{wu2021self}
Wu, L., Lin, H., Gao, Z., Tan, C., Li, S., et~al.: Self-supervised on graphs:
  Contrastive, generative, or predictive. arXiv preprint arXiv:2105.07342
  (2021)

\bibitem{wu2020comprehensive}
Wu, Z., Pan, S., Chen, F., Long, G., Zhang, C., Philip, S.Y.: A comprehensive
  survey on graph neural networks. IEEE transactions on neural networks and
  learning systems  \textbf{32}(1),  4--24 (2020)

\bibitem{xu2021self}
Xu, M., Wang, H., Ni, B., Guo, H., Tang, J.: Self-supervised graph-level
  representation learning with local and global structure. arXiv preprint
  arXiv:2106.04113  (2021)

\bibitem{yin2022autogcl}
Yin, Y., Wang, Q., Huang, S., Xiong, H., Zhang, X.: Autogcl: Automated graph
  contrastive learning via learnable view generators. In: Proceedings of the
  AAAI Conference on Artificial Intelligence. vol.~36, pp. 8892--8900 (2022)

\bibitem{you2021graph}
You, Y., Chen, T., Shen, Y., Wang, Z.: Graph contrastive learning automated.
  arXiv preprint arXiv:2106.07594  (2021)

\bibitem{you2020graph}
You, Y., Chen, T., Sui, Y., Chen, T., Wang, Z., Shen, Y.: Graph contrastive
  learning with augmentations. Advances in Neural Information Processing
  Systems  \textbf{33} (2020)

\bibitem{zeng2019graphsaint}
Zeng, H., Zhou, H., Srivastava, A., Kannan, R., Prasanna, V.: Graphsaint: Graph
  sampling based inductive learning method. arXiv preprint arXiv:1907.04931
  (2019)

\bibitem{zhang2018link}
Zhang, M., Chen, Y.: Link prediction based on graph neural networks. Advances
  in Neural Information Processing Systems  \textbf{31},  5165--5175 (2018)

\bibitem{zhao2021graph}
Zhao, H., Yang, X., Wang, Z., Yang, E., Deng, C.: Graph debiased contrastive
  learning with joint representation clustering. IJCAI (2021)

\bibitem{zhu2020self}
Zhu, Q., Du, B., Yan, P.: Self-supervised training of graph convolutional
  networks. arXiv preprint arXiv:2006.02380  (2020)

\bibitem{zhu2020deep}
Zhu, Y., Xu, Y., Yu, F., Liu, Q., Wu, S., Wang, L.: Deep graph contrastive
  representation learning. arXiv preprint arXiv:2006.04131  (2020)

\bibitem{zhu2021graph}
Zhu, Y., Xu, Y., Yu, F., Liu, Q., Wu, S., Wang, L.: Graph contrastive learning
  with adaptive augmentation. In: Proceedings of the Web Conference 2021. pp.
  2069--2080 (2021)

\bibitem{zitnik2017predicting}
Zitnik, M., Leskovec, J.: Predicting multicellular function through multi-layer
  tissue networks. Bioinformatics  \textbf{33}(14),  i190--i198 (2017)

\end{thebibliography}
\end{document}